\documentclass[11pt,a4paper]{article}
\usepackage{authblk}
\usepackage[hyperref]{acl2020}
\usepackage{times}
\usepackage{url}
\usepackage{latexsym}
\usepackage{subfiles}
\hypersetup{ breaklinks=true, colorlinks=true}

\usepackage{microtype}

\aclfinalcopy % Uncomment this line for the final submission

\title{Natural Answer Generation: From Factoid Answer to Full-length Answer using Grammar Correction}

\author[1]{\textbf{Manas Jain}}
\author[2]{\textbf{Sriparna Saha}}
\author[1]{\textbf{Pushpak Bhattacharyya}}
\author[3]{\textbf{Gladvin Chinnadurai}}
\author[3]{\textbf{Manish Kumar Vatsa}}
\affil[1]{Indian Institute of Technology Bombay, Mumbai}
\affil[2]{Indian Institute of Technology Patna}
\affil[3]{LG Soft India}
\affil[1]{\textit{\{manasjain, pb\}@cse.iitb.ac.in}}
\affil[2]{\textit{sriparna@iitp.ac.in}}
\affil[3]{\textit{\{gladvin.durai, manish.vatsa\}@lge.com}}

\date{}

\begin{document}
\maketitle
% \begin{abstract}
% In Question Answering domain, to present the user with a more conversational experience the task of generation of "full length answer" from factoid answer becomes very important. In recent years, the task of Question Answering over passages (reading comprehension) has evolved into a very active research area. A reading comprehension system extracts a span of text, consisting of named entities etc., which serve as the answer to a given question (known as "factoid answer"). However, these spans of text would result in an unnatural reading experience to user in systems like chatbots and speech assistants. Usually, dialogue systems solve this issue by using template-based language generation. These systems, though adequate for a domain-specific task, are too restrictive and predefined for a domain-independent system. Our system outputs a full-length answer given as input a question and the extracted factoid answer. In this work we present an unsupervised rule-based approach built on using constituency and dependency parse trees of questions with transformer based Grammar Error Correction model GECToR (2020) as a post processing step. Comparison was made with 2 supervised approaches (i) Modified Pointer Generator (SOTA) (ii) Fine-tuned DialoGPT for factoid questions and existential (yes-no) questions. The developed model generates answers that are more accurate and fluent than the supervised approaches. There is an improvement in ROUGE-1 score and inference time is reduced by 85\% as compared to the state of art model.
% \end{abstract}
\begin{abstract}
    Question Answering systems these days typically use template-based language generation. Though adequate for a domain-specific task, these systems are too restrictive and predefined for domain-independent systems. This paper proposes a system that outputs a full-length answer given a question and the extracted factoid answer (short spans such as named entities) as the input. Our system uses constituency and dependency parse trees of questions. A transformer-based Grammar Error Correction model GECToR (2020), is used as a post-processing step for better fluency. We compare our system with (i) Modified Pointer Generator (SOTA) and (ii) Fine-tuned DialoGPT for factoid questions. We also tested our approach on existential (yes-no) questions with better results. Our model generates accurate and fluent answers than the state-of-the-art (SOTA) approaches. The evaluation is done on NewsQA and SqUAD datasets with an increment of 0.4 and 0.9 percentage points in ROUGE-1 score respectively. Also the inference time is reduced by 85\% as compared to the SOTA. The improved datasets used for our evaluation will be released as part of the research contribution.
\end{abstract}

\section{Introduction}
Question answering (QA) is an exercise of finding solutions for a query from a given paragraph. Normally small spans of
text, inclusive of named entities, dates, etc. are extracted as answers. However, knowledge-base (KB) orientated QA systems extract factoid solutions by using a structured query or
neural representation of the question. As a natural extension and post-processing step, the retrieved factoid answer is transformed into a full-length natural sentence.
Unlike conversational chat-bots designed to mimic human communique with out worrying to be factually correct, or assignment-orientated dialogue system which places the retrieved solution in a predefined template, our approach routinely generates correct full-length solutions, thereby, improving it’s utilization in these  situations.\\
\\
\textit{\textbf{Question} : Who was the duke in the battle of hastings ?}\\
\textit{\textbf{Factoid answer} : william the conqueror}\\
% \textit{\textbf{RBV2 output} : lewis partnership begin started as a single shop on oxford street in london, opened in 1864 by john.}\\\
\textit{\textbf{Target} :  [The duke in the battle of hastings was william the conqueror. , William the conqueror was the duke in the battle of hastings.]}\\
\\
\centerline{\textit{Example 1 - Sample from SqUAD dataset}}\\
\\
% \begin{table}[]
%     \centering
%     \textit{\textbf{Question} : Who was the duke in the battle of hastings ?}\\\
% \textit{\textbf{Factoid answer} : william the conqueror}\\\
% % \textit{\textbf{RBV2 output} : lewis partnership begin started as a single shop on oxford street in london, opened in 1864 by john.}\\\
% \textit{\textbf{Target} :  [The duke in the battle of hastings was william the conqueror. , William the conqueror was the duke in the battle of hastings.]}
%     \caption{Example 1 - Sample from SqUAD dataset}
%     \label{exam:1}
% \end{table}
Our overall research contributions are listed as
follows:
\begin{enumerate}
    \item We achieve superior performance by incorporating a pre-trained transformer encoder
    GEC sequence tagging system as a post processing step in our rule based approach. In our experiments,
    encoders from RoBERTa outperform
    three other cutting-edge transformer encoders
    (XLNet, BERT).
    % \item BLEU and ROUGE scores are not very good evaluation metrics for this task of natural answer generation from factoid answers. This is because since most part of the answer is directly copied from the question and factoid answer so the scores are not representative of the errors the model is making.
    \item We present a rule based approach for existential questions (Yes/No questions) where Yes/No is considered as the factoid answer and natural answer is generated by rearranging noun phrase and verb phrase present in the question. We achieve good metrics (BLEU, ROUGE-1,2,L) and also analyze results of using Grammar correction model, GECTOR, on top of the developed rule based system. 
    \item We have made the existing dataset for this task more accurate by correcting grammar errors in GOLD answer and have added alternate answers wherever necessary. We also have created a small dataset for Existential QA having different types of indirect questions as well. We will open source all the improved datasets for further research.

\end{enumerate}
The rest of the paper is organised as follows: Firstly we discuss some recent works and related literature in section \ref{sec:related-work}, after which we give details about the data used for evaluating our system in section \ref{sec:data}. After that we talk about our approach in section \ref{sec:approach}; rule based in section \ref{sec:rule-based} (factoid questions in section \ref{sec:rule-based-factoid}, existential questions in section \ref{sec:rule-based-yesno}) and fine-tuned DialoGPT in section \ref{sec:dialogpt}. Following up on this, we provide details about our experimental setup and talk about the GCM used as a post processing step in section \ref{sec:experimental-setup}. Then in section \ref{sec:results} we provide the results \& evaluation of our approach; compare performance from other approaches. Then in section \ref{sec:error-analysis}, we give extensive error analysis of all approaches (Modified Pointer Generator [SOTA] in section \ref{sec:error-analysis-mpg}, fine-tuned DialoGPT in section \ref{sec:error-analysis-dgpt} and rule based in section \ref{sec:error-analysis-rbv2}) presented in the paper and discuss some ways to overcome them. Lastly, we conclude our paper discussing future work in section \ref{sec:conclusion}.

\section{Related Work}
\label{sec:related-work}
% papers to be covered of topics Knowledge base generative QA, Natural answering in knowledge base, conversational QA systems and dialog systems.
The task of retrieving  exact information nowadays has become a very complex and time consuming process as very huge amount of raw information are freely available. This situation
has prompted in development of QA systems, to directly return the correct answer to a question
asked in natural language without making search and the information filtering tasks, as it is often the case with search engines.\\
We generally notice two forms of representations
addressed in the literature. The answer can take the
form of a paragraph selected from a set of text
passages retrieved from the web \cite{DBLP:journals/corr/abs-1809-03275},
\cite{du-cardie-2018-harvesting}, \cite{DBLP:journals/corr/WangJ16a}, \cite{DBLP:journals/corr/WangYT17}, \cite{Oh_Torisawa_Hashimoto_Iida_Tanaka_Kloetzer_2016}, or it can be
the exact answer to the question extracted from a
knowledge base or a span of texts from a reading comprehension paragraph.\\
Despite the abundance of works in the field of
QA systems, the answer generation issue has received little attention. A first approach indirectly addressing
this task has been proposed in \cite{brill-etal-2002-analysis}. Indeed, the authors aimed at diversifying
the possible answer patterns by permuting the question’s words in order to maximise the number of
retrieved documents that may contain the answer
to the given question.
Another answer representation approach based on rephrasing rules has also
been proposed in \cite{10.1145/336597.336644} within the context of
query expansion task for document retrieval and
not purposely for the question-answering task.\\
It should be noted that these approaches only
intend to diversify as much as possible the answer
representation patterns to a given question in order
to increase the probability of extracting the correct
answer from the Web and do not focus on the answer’s representation itself. It should also be noted
that these approaches are only applicable for QA systems
which extract answers as a text snippet and cannot be applied to short answers usually extracted
from knowledge bases. \\
Some recent works are presented in \cite{pal-etal-2019-answering} and \cite{DBLP:conf/inlg/AkermiHH20}. Former work tried to tackle this issue by proposing a supervised approach based on modifying pointer generator network \cite{see-etal-2017-get} while the latter proposed a transformer based unsupervised approach incorporating the use of language models to evaluate different possible answer structures. In \cite{pal-etal-2019-answering}, the model was trained on a
small data set whose question/answer pairs were
extracted from machine comprehension datasets
and augmented manually which make generalization and capturing variation very limited. In \cite{DBLP:conf/inlg/AkermiHH20}, authors have used syntactic parser to form rules to get fragments useful for the formation of natural answer. They assumed that only one word could be missing and it should be located before the factoid answer within the identified structure. This assumption cannot be generalized and can lead to incomplete answers with grammatical errors.\\
Our answer generation approach differs from
these works as it is completely rule based. We have utilized \cite{omelianchuk-etal-2020-gector} by which any number of words at any place can be added or deleted. Indeed, we build upon the intuitive hypothesis that a full length can be made by reformulation of the words given in the question and factoid answer with few insertion/deletion in between.

\section{Data}
\label{sec:data}

There is just one available dataset \cite{pal-etal-2019-answering} for this task  created from reading comprehension dataset having 15000 manually annotated, 300000 automatically annotated from SQuAD \cite{rajpurkar-etal-2016-squad}, HarvertingQA \cite{du-cardie-2018-harvesting} and 420 data points in test dataset taken from NewsQA \cite{trischler-etal-2017-newsqa}. After going through the dataset, we realized the available dataset is not of high quality, having numerous grammatically incorrect questions/answers and also wrong or grammatically incorrect target answer in many cases. Due to this, improving the quality of dataset is the need of the hour. \\
In natural language generation (NLG) systems, there can be more than one correct answer but that is not incorporated well in the available dataset.
\\
\\
\textit{\textbf{Question} : Who is the ceo of google ?}\\
\textit{\textbf{Factoid answer} : Sundar Pichai}\\
% \textit{\textbf{RBV2 output} : lewis partnership begin started as a single shop on oxford street in london, opened in 1864 by john.}\\\
\textit{\textbf{Target} :  [(i) Sundar Pichai is the ceo of google. (ii) The ceo of google is Sundar Pichai.]}\\
\\
% \textit{\textbf{Q-? \textbf{FA}-Sundar Pichai \textbf{A1}-Sundar Pichai is the ceo of google. \textbf{A2}-The ceo of google is Sundar Pichai.}}
% \\
% \centerline{\textit{Example 1 - Sample from "X" dataset}}\\
% \\
In the existing dataset we see only target (i) type annotations but target (ii) is also correct way to answer this question and should be added in the annotation. So we improve the quality of the available dataset to handle the above mentioned issues. We sampled 7200 data points from 15000 manually annotated SqUAD samples \cite{pal-etal-2019-answering}, 420 data points from NewsQA \cite{pal-etal-2019-answering} and made the required changes in target answers; some data points were removed due to incomplete question/answer. As given in Table \ref{tab:dataset-created}, our improved dataset has 6768 data points from SQuAD and 380 data points from NewsQA.
% which were used to compare different models as given in table \ref{tab:squad} and \ref{tab:freq} 
We have also created 166 data points of existential QA dataset containing different varieties and forms of asking questions, including indirect questions. The codes and the data sets will be publicly available after the acceptance of the paper. 
%We will open source the improved data for further research.\\
\\
\\
\textit{\textbf{Question - type (i)} : Does my fridge support quick freeze feature?}\\
\textit{\textbf{Question - type (ii)} : Can you tell me if my fridge supports quick freeze feature?}\\
\textit{\textbf{Target} :  [ No, your fridge does not support quick freeze feature. OR Yes, your fridge supports quick freeze feature.]}\\
\\
\centerline{\textit{Example 2 - Sample from Yes/No dataset}}\\
\begin{table}[]
  \centering
  
  \begin{tabular}{lc}
    \hline
    Dataset&Count\\
    \hline
    % mpg\cite{pal-etal-2019-answering}&64.1&85.7&72.5&78.8\\
    NewsQA (Factoid) & 380\\
    SqUAD (Factoid) & 6768\\
    Yes/No (Existential) & 166\\
    \hline
    \end{tabular}
\caption{\label{tab:dataset-created} Dataset used for our evaluation}
\end{table}
\section{Approach}
\label{sec:approach}
In this section we explain the rule based approach and fine-tuned DialoGPT approach developed.
\subsection{Rule Based Approach}
\label{sec:rule-based}
\subsubsection{Factoid Questions}
\label{sec:rule-based-factoid}
After observing a large number of examples in the available dataset we were able to find patterns in the formation of natural answer using the sentence structure of the question at its core. Initially the idea was to check the accuracy by just replacing the \textit{WH} words present in the question with the factoid answer; we refer that approach as Rule Based V1 in below examples. After analysing the output of the above idea on the failed cases led to a finding of pattern related to the position of auxiliary verb and the main verb. We used the constituency and dependency parsing output of the question to find positions of auxiliary verb, main verb, noun phrases and verb phrases present in the question and designed the algorithm; we refer this improved version of our approach as Rule Based V2 (RBV2). Outputs of constituency parser with Elmo Embeddings given in \cite{Joshi2018ExtendingAP} and deep biaffine attention neural dependency parser \cite{Dozat2017DeepBA} were extensively used in the algorithm developed. We used open source AllenNLP library \cite{Gardner2017AllenNLP} APIs of the above 2 parsers in developing our rule based system. \\
Below we will explain our approach using some examples and also discuss implementation details.
In the first version of our rule based approach (Rule Based V1), we have just replaced the \textit{WH} words (\textit{what, when, why, who, how etc.}) present in the question with the factoid answer. The \textit{WH} word in the question was found by using the outputs of POS tags of the AllenNLP constituency parser \cite{Joshi2018ExtendingAP}. If the tag is "WP" or "WRB" or "WDT" then we replace that word with factoid answer. This phenomenon where sentence topic appears at the front of the sentence as opposed to in a canonical position further to the right is known as topicalization \cite{Prince1998OnTL}. Some examples are stated below for better understanding of the approach:- \\
\\
\\
\textit{\textbf{Question} : What is the capital of India?}\\
\textit{\textbf{Factoid answer} : Delhi}\\
\textit{\textbf{Rule Based V1} : Delhi is the capital of India}\\
\textit{\textbf{Target answer} : Delhi is the capital of India}\\
\\
\centerline{\textit{Example 3 - Self made Sample}}\\
\\
\textit{\textbf{Question} : what was the space station crew forced to take shelter from?}\\
\textit{\textbf{Factoid answer} : a piece of debris}\\
\textit{\textbf{Rule Based V1} : a piece of debris was the space station crew forced to take shelter from}\\
\textit{\textbf{Target answer} : the space station crew was forced to take shelter from a piece of debris}\\
\\
\centerline{\textit{Example 4 - Sample from NewsQA dataset}}\\
\\
In the second version (Rule Based V2[RBV2]), we modify the above approach based on the position of auxiliary verb and main verb present in the question. We formulate the algorithm as to solve the problem of ordering of natural answer, i.e., answer followed by portion from question or portion of question followed by answer. So, if the main verb and auxiliary verb are consecutive, factoid answer appears in the starting otherwise we add factoid answer in the end. In the latter case, we start our answer from the word after the auxiliary verb, till the main verb is encountered, then the auxiliary word is added to the answer string. Then we copy the part of question after the main verb, finally adding the factoid answer. \\
If question does not have verb in it then we add all words after auxiliary word present in the question to our answer. We then add auxiliary verb, finally adding the factoid answer. Some sample example outputs using this approach are stated below:-\\
 \\
% \textbf{CASE :- Main Verb not present}\\
\textit{\textbf{Question} : What is the capital of India?}\\
\textit{\textbf{Factoid answer} : Delhi}\\
\textit{\textbf{Rule Based V2(RBV2)} : the capital of India is Delhi}\\
\\
\centerline{\textit{\textbf{CASE :- Main Verb not present}}}\\
\\
% \\\
% \textbf{CASE :- Auxiliary Verb and Main Verb not together}\\
\textit{\textbf{Question} : what was the space station crew forced to take shelter from?}\\
\textit{\textbf{Factoid answer} : a piece of debris}\\
\textit{\textbf{Rule Based V2(RBV2} : the space station crew was forced to take shelter from a piece of debris}
\\
\\
\textit{\textbf{CASE :- Auxiliary Verb and Main Verb not \centerline{together}}}\\
\subsubsection{Existential Questions (Yes/No)}
\label{sec:rule-based-yesno}
It would be incomplete if we just limit this task of natural answering to just factoid questions. This task can have importance in the existential question type as well in systems or apps tackling user queries using speech assistants or chatbots. So, we tried formulating a rule based approach for existential questions or yes/no questions using the dependency and constituency parse tree of the questions. Generally such questions have a common structure: auxiliary verb (AUX) followed by a noun phrase (NP) and then verb phrase (VP) in the end, i.e., AUX-NP-VP. The natural answers to such questions can be made by reordering the above parts to NP-AUX-VP. This was implemented using the output of AllenNLP dependency parse tree model. In addition we start the answer with "yes," or "no," so as to create a more natural sounding answer. \\
\\
\textit{\textbf{Question} : Can you tell if fridge supports quick freeze feature?}\\\
\textit{\textbf{Factoid answer} : Yes}\\
% \textit{\textbf{RBV2 output} : lewis partnership begin started as a single shop on oxford street in london, opened in 1864 by john.}\\\
\textit{\textbf{RB} :  Yes, fridge does \textcolor{red}{supports} quick freeze feature.}\\
\textit{\textbf{RB + RoBERTa} :{Yes, fridge does \textcolor{green}{support} quick freeze feature.}}\\
\\
\centerline{\textit{Example 5 - Sample from Yes/No dataset}}\\

\subsection{Fine-tuned DailoGPT} 
\label{sec:dialogpt}
In order to resolve the problem of fluency which is very important for the task of natural answering, we explored the latest cutting edge GPT \cite{radford2019language} models which are very popular in generating fluent responses. We found the DialoGPT \cite{zhang-etal-2020-dialogpt} model is the best fit for this task as it was trained on large corpus of dialogue conversations which is somewhat relevant to the task in hand. We fine tuned the above pre-trained model publicly available on approximately 13000 manually annotated data points given by \cite{pal-etal-2019-answering}.\\  
DialoGPT (dialogue generative pre-trained transformer) \citep{zhang-etal-2020-dialogpt} is a tun-able giga word scale neural network model for generation of conversational responses, trained on Reddit data.
Trained on 147M conversation-like exchanges
extracted from Reddit comment chains over
a period spanning from 2005 through 2017,
DialoGPT extends the Hugging Face PyTorch
transformer to attain a performance close to
human both in terms of automatic and human
evaluation in single-turn dialogue settings.
Our aim was to develop a system which produces natural answer given any type of question and factoid answer in the input as DialoGPT was created with the aim of neural response generation and the development of more intelligent open domain dialogue systems. \\
DialoGPT extends GPT-2 \citep{radford2019language} to address the challenges of conversational neural response generation. Neural response generation is a subcategory of text-generation that shares the objective of
generating natural-looking text (distinct from any
training instance) that is relevant to the prompt.
Normally DialoGPT model was created to make conversational chatbots and their fine-tuning is also done for building  conversational agent where the input is the question asked and all the previous dialogues are kept as series of context and are passed as input to the model for training. But here for our task, we concatenate the question with its extracted factoid answer and keep manually annotated answers as target in fine-tuning the model. For inference, question and factoid answer are concatenated and provided as input to the fine-tuned model to generate a response.

\section{Experimental Setup}
\label{sec:experimental-setup}
 We have used Tesla T4 16GB GPU to carry out the experiments. For factoid questions, we have used two datasets having 380 and 6768 data points as given in Table \ref{tab:dataset-created}. Experimental results are shown in Table \ref{tab:freq} and \ref{tab:squad}, respectively. For existential questions, we have used created data set with 166 examples. Results of confirmatory dataset are reported in Table \ref{tab:yes-no}.\\
As a post processing step of all our rule based approaches, we have used a pre-trained transformer encoder, grammar error correction (GEC) given in \cite{omelianchuk-etal-2020-gector}. This model was available with 3 cutting edge transformer encoders namely BERT, RoBERTa and XLNet. Experiments were carried out using all 3 above encoder based GEC models as post processing steps in our rule based approach.\\
For fine-tuning DialoGPT, we took a pretrained DialoGPT-small (117M parameters) and fine-tuned with around 13000 manually annotated samples data from \cite{pal-etal-2019-answering}. We trained the model for 8 epochs.
% {\bf IS IT CORRECT?? DialoGPT is fine-tuned using around 13000 manually annotated samples and it is finetuned for 8 epochs.}
The results on 380 data points (cross validation) of NewsQA dataset by the fine-tuned model are reported in Table \ref{tab:freq}. \\

\section{Results}
\label{sec:results}
We use standard BLEU \cite{papineni-etal-2002-bleu} (NLTK), ROUGE-1, 2, L \cite{lin-2004-rouge} (rouge-score) metrics to evaluate our system and compare our system with other approaches.
In Table \ref{tab:freq}, \ref{tab:squad}, \ref{tab:yes-no} : "RBV2+RoBERTa" means our rule based approach with grammar correction performed by RoBERTa encoder and so on.\\
Table \ref{tab:example} illustrates a qualitative comparison of outputs from different approaches explored in this paper.\\
In Table \ref{tab:freq}, we see an increase in BLEU, ROUGE-2, ROUGE-L scores on using RoBERTa encoder Grammar Correction Model (GCM) as compared to not using it. It is also clear that RoBERTa based encoder GCM is superior as compared to other encoders due to higher BLEU and ROUGE scores. Our developed approach attains very comparable results in terms of BLEU and ROUGE-1, 2, L scores and reduces inference time by 85\% as compared to the state of the art MPG model. Avg. time in table \ref{tab:freq}, \ref{tab:squad} denotes the average time taken by the model or algorithm to generate answer for 1 (question, factoid answer) input. ROUGE-1 and ROUGE-L scores are almost same with a difference of 3 and 1 percentage points in BLEU and ROUGE-2 scores, respectively. BLEU and ROUGE scores provided in all the tables are on the scale of 100.\\
In Table \ref{tab:squad}, reported ROUGE-1 and ROUGE-L scores are almost same. BLEU and ROUGE-2 scores for our approach (RBV2 + GCM) are a bit lesser than SOTA model (MPG).\\
% {\bf DID NOT GET There is some difference in the BLEU and ROUGE-2 scores, our developed system on the lower side as compared to SOTA model.}\\
There are instances in the above tables where employing a GCM sometimes reduces the BLEU or ROUGE scores especially in Table \ref{tab:squad}. This phenomenon is very much related to the target (GOLD) answers based on which the scores are calculated. This can occur because of insertion/deletion of punctuation in between by GCM but not present in target answer and vice-versa. In many cases, target answers do not follow correct grammar which sometimes leads to lower scores. But in such cases also the overall quality, fluency and adequacy of the answers improved by GCM are much better.\\
 Table \ref{tab:freq} illustrates that performance of fine-tuned DialoGPT is comparatively very low as compared to other approaches in cross evaluation. Main problem with this approach was the problem of hallucination as explained in \cite{maynez-etal-2020-faithfulness} which decreases the accuracy of the approach and hence we conclude that it is not useful for this task. Due to that, we have skipped the results of the fine-Tuned model in Table \ref{tab:squad}.\\
In Table \ref{tab:yes-no}, scores are calculated on a very small dataset and best scores are achieved by  simply employing the rule based model without using GCM. We still argue to use a GCM as a post processing step in this type as well due to its ability to improve the overall quality of the answers. This improvement in quality can not be measured using these scores but can surely improve the user satisfaction. This kind of task in existential questions is in best of our knowledge first time presented so there is no baseline model to compare our results with. 
% \begin{itemize}
%     \item \textbf{How the input was provided in finetuning the DGPT model?}\\\
%     Normally DIALOGPT models are used to make conversational chatbots and their finetuning is also done for making conversational agent where the input is the question asked and all the previous dialogues are kept as series of context and are passed as input to the model for training.
% \end{itemize}

% Here in our task since all our questions are independent we
% have passed the input as question and the factoid answer as context and response as the GOLD standard answer that we have. Here instead of a series of conversations passed as context normally in our case we only give the factoid answer as a context to the model.  
% \subsection{Factoid questions}
% Handling models like GPT-2 and DIALOGPT which are trained on huge corpus of data is very difficult. Finetuning these models is even more difficult as these models require a very available memory to run on the machine. We tried different combinations of data for fine-tuning like combining manually annotated data with the auto annotated examples \cite{pal-etal-2019-answering} etc. But finally best results were achieved when only the manually annotated data was used to fine-tune the DIALOGPT model.

\begin{table*}
  \centering
  \begin{tabular}{lccccc}
    \hline
    Model&BLEU& ROUGE-1&ROUGE-2&ROUGE-L&Avg. time (sec.)\\
    \hline
    MPG\citeyearpar{pal-etal-2019-answering}&84.9&95.7&89.4&93.9&2.54\\
    RBV2 & 79.1 & 96.1 & 85.5 & 93.1 & 0.382\\
    RBV2+BERT & 77.6 & 94.4 & 85.4 & 92.4 & 0.397\\
    RBV2+RoBERTa & 81.7 & 95.7 & 88.2 & 93.6 & 0.394\\
    RBV2+XLNET & 80.3 & 94.8 & 87.0 & 92.9 & 0.4\\
    DialoGPT & 50.3 & 73.4 & 49.3 & 70.0 & 0.908\\
    \hline
\end{tabular}
\caption{\label{tab:freq} Results on 380 data points of NewsQA dataset}
\end{table*}

\begin{table*}
  
  \centering
  \begin{tabular}{lccccc}
    
    \hline Model&BLEU& ROUGE-1&ROUGE-2&ROUGE-L&Avg. time (sec.)\\ \hline
    
    MPG\citeyearpar{pal-etal-2019-answering}&75.8&94.4&87.4&91.6 & 2.54\\
    RBV2 & 74.8 & 95.3 & 83.1 & 90.3 & 0.399 \\
    RBV2+BERT & 71.5 & 93.9 & 82.4 & 89.5 & 0.411\\
    RBV2+RoBERTa & 72.1 & 94.0 & 83.1 & 89.8 & 0.411\\
    RBV2+XLNET & 71.2 & 93.6 & 82.3 & 89.4 & 0.413\\
    % DialoGPT & 50.3 & 73.4 & 49.3 & 70.0\\
    \hline
\end{tabular}
\caption{\label{tab:squad} Results on 6768 data points of SqUAD dataset}
\end{table*}

\begin{table}
  \centering
  
  \begin{tabular}{lcccc}
    \hline
    Model&BLEU& R-1&R-2&R-L\\
    \hline
    % mpg\cite{pal-etal-2019-answering}&64.1&85.7&72.5&78.8\\
    RB & 70.2 & 87.3 & 75.0 & 84.8\\
    RB+BERT & 62.7 & 85.5 & 71.6 & 83.4\\
    RB+RoBERTa & 66.6 & 84.5 & 73.0 & 84.2\\
    RB+XLNET & 67.5 & 86.6 & 74.0 & 84.6\\
    % DialoGPT & 50.3 & 73.4 & 49.3 & 70.0\\
    \hline
    \end{tabular}
\caption{\label{tab:yes-no} Results on 166 data points of existential questions dataset created by us; Here in the table R represents Rouge, R-1 means ROUGE-1 and so on}
\end{table}

\begin{table*}
\centering
\begin{tabular}{ll}
\hline
\textbf{Input} & \textbf{Output} \\
\hline
\textbf{Ques} - where was the bus going ? & \textbf{MPG} \cite{pal-etal-2019-answering} - the bus \textcolor{red}{going was at} phoenix, arizona. \\
\textbf{Factoid Ans.} - phoenix, arizona & \textbf{FT DialoGPT} [ours] - the bus was going to phoenix, \textcolor{red}{anrizona}. \\ & \textbf{RBV2} [ours] - the bus \textcolor{green}{was going} phoenix, arizona . \\ & \textbf{RBV2+GCM} [ours] - The bus was going \textcolor{green}{to} Phoenix, Arizona. \\ 
\hline
\end{tabular}
\caption{\label{tab:example}
Comparison of outputs from all approaches discussed in the paper for an input example. Here MPG represents the state of the art deep learning model using Pointer Generator technique. FT DialoGPT represents the results of the fine-tuned model of DialoGPT for this task. RBV2+GCM represents the results of using the GEC Model as a post processing step. Here we used the RoBERTa encoder GECTOR model as GCM. 
}
\end{table*}

\section{Error Analysis}
\label{sec:error-analysis}
Below we present some qualitative discussion and error analysis of answers generated by existing approaches and our proposed approach.  
\subsection{Modified Pointer Generator(MPG)}
\label{sec:error-analysis-mpg}
This approach was taken from \cite{pal-etal-2019-answering}.
The main limitations of this approach are stated in below points. Also there were failure cases wherein the model just outputs the question itself which may be due to model becomes biased towards adding more part from the question than the factoid answer which results in complete copying of the question in some cases. Below are the main types of failure cases stated :- 
\begin{itemize}
    \item Incoherent sentence due to failure in reasoning
    \item Repetition of words
    \item Outputs only the factoid answer
    \item Outputs clausal answers
    \item Failure to incorporate morphological variations
\end{itemize}
This can also be seen in Table \ref{tab:example} where MPG makes error in answer generation. Word positions of was and going are interchanged and "at" is added which is wrong, correct addition should be "to".\\
Overall, this model doesn't attain good results even for very straight forward example cases present in our dataset and so using it for general case queries would not be very beneficial. Also inference time of this model is very high (last column of Table \ref{tab:freq},\ref{tab:squad}).
\subsection{Fine-tuned DialoGPT }
\label{sec:error-analysis-dgpt}
Problem of adding unwanted things in the final answers which doesn’t have any mention in the question and the factoid answer often called as hallucination \cite{maynez-etal-2020-faithfulness} is the main shortcoming of this model.\\
There are instances where factoid answer is not even present in the final answer. Also there are numerous cases where DialoGPT model makes error in copying numerical data for \textit{e.g.} year, number etc.
The model has some errors copying the proper nouns as given in the questions. The final answer has those names but with changed spelling. (\textit{e.g.}:- elizabeth - elizabetha; alexander - alexandrick). This is also evident from the example given in Table \ref{tab:example} where DialoGPT has changed arizona spelling to "anrizona". This leads to low BLEU and ROUGE scores.
For eg,
\\
\\
\textit{\textbf{Question} : What is going live on tuesday?}\\
\textit{\textbf{Factoid answer} : web-based on-demand television and movie service}\\
\textit{\textbf{Fine-Tuned DialoGPT} : on tuesday, the web-based version of "net based" television and film service.}\\
\textit{\textbf{Target answer} :  web-based on-demand television and movie service is going live on tuesday.}\\
\\
\centerline{\textit{Example 6 - Sample from NewsQA dataset}}\\
\\
In the above example we find very poor quality of answer generated. Here we see additional "net-based" getting added which makes this model unreliable for this task.\\
% Also DialoGPT model has high bias of generating related things in the natural answer that is why BLEU and ROUGE scores are very less compared to other approaches.
\subsection{Rule Based Model}
\label{sec:error-analysis-rbv2}
This approach works by reordering question sentence structure and copy pasting the factoid answer; and so if factoid answer is not factual based or is a clausal answer then this approach may fail. Also the generated answers may be grammatically wrong in terms of missing a word like in, is, to etc. which is corrected by the transformer based grammar correction used as a post processing step; other type of grammatical error by rule based approach is incorrect positioning of AUX word (\textit{e.g.} is, are etc) in the answer which is not corrected by the \cite{omelianchuk-etal-2020-gector} sometimes.
\\
\\
\textit{\textbf{Question} : where did lewis partnership begin?}\\
\textit{\textbf{Factoid answer} : started as a single shop on oxford street in london, opened in 1864 by john.}\\
\textit{\textbf{RBV2} : lewis partnership begin started as a single shop on oxford street in london, opened in 1864 by john.}\\
\textit{\textbf{Target answer} :  lewis partnership begin started as a single shop on oxford street in london, opened in 1864 by john.}\\
\\
\centerline{\textit{Example 7 - Sample from SqUAD dataset}}\\
\\
In the above example, output answer had both begin and started in it which is not right, this is because the factoid answer contains a clause having verb part included. Currently, in our system we are not checking the factoid answer structure to define our answers and hence for these examples this model may fail.
Since the approach works on the question structure so if question is not properly well formed or incomplete then the answers will not be correct.
Instances where the question is of type "how many"; the word "many" can be added or not added based on the type of factoid answer given. In such cases we rely on the GCM model to perform necessary corrections but sometimes the GCM model fails to make the changes.\\
Questions having a subordinate clause are a challenge to this approach. Such examples generally have 2 WH words and so sometimes are difficult to handle. With some modifications we will be able to handle those questions as well in our rule based approach by first finding out the main clause in the question and masking the subordinate clause temporarily considering if that subordinate clause never existed and then unmasking it after answer generation. \\
As highlighted by \citet{van-miltenburg-etal-2021-underreporting}, the under-reporting of errors and lack of extensive error analysis of NLG system output is quite common nowadays. This prevents  researchers to get an idea about the specific weakeness of SOTA and improved model. So in this work we categorised the errors for the 6768 data points of SqUAD dataset. These
errors are categorised as: extra words like do, does, is, was; incorrect words like much, many; misplaced words like is, were, was, are, has; missing words like in, to, on, during, by, until, through, at, after, between; wrong preposition, word order. Count of these categories are reported in \ref{tab:grammar-errors}. The GCM as the post processing step in our approach is able to correct most of above errors for our system and thus improving the quality of our generated answers as can be seen in Table \ref{tab:example}.
\begin{table}
  \centering
  
  \begin{tabular}{lc}
    \hline
    Grammar Error&Count\\
    \hline
    % mpg\cite{pal-etal-2019-answering}&64.1&85.7&72.5&78.8\\
    Grammar Error [extra] & 103\\
    Grammar Error [incorrect] & 25\\
    Grammar Error [misplaced] & 254\\
    Grammar Error [missing] & 815\\
    \hline
    \end{tabular}
\caption{\label{tab:grammar-errors} Count of categories of grammar errors by the rule based algorithm without using the GCM. These numbers are for the 6768 data points from SqUAD dataset}
\end{table}
\section{Conclusion \& Future Work}
\label{sec:conclusion}
In this work, we have worked on the task of generating full-length natural answers given the question and the factoid answer. We have solved this task by designing a rule based approach using syntactic parser. A Grammar Correction Model (GCM) is used as a post processing step to improve the fluency of generated natural answer. Our approach RBV2 and RoBERTa based encoder GCM achieves superior results
than the state of the art deep learning model in terms of ROUGE-1 score, quality of the answers generated and inference time. This system can be used at the final stage of any domain-specific QA system or answering user troubleshooting queries where factoid answer is extracted by a knowledge base or context paragraphs. This approach is developed using general rules of answer generation and so can be applied to all domains as compared to supervised system which gets biased to the type of training data given. We have also improved the quality of existing dataset by creating 2 sets having 6768 and 380 data points, respectively. We have also created a dataset of 166 data points of existential (yes/no) questions.\\
We plan to make our system more robust especially for questions having subordinate clauses present. We will work on making a complete system which can classify existential and factoid questions and use our developed system on top of that. We plan to give our generated answers for review to some proficient English speaker and ask for scores on fluency, adequacy of our generated answer and other approachs' answers. Further work needs to be done to investigate the performance of reinforcement learning based techniques for solving this task, keeping BLEU or ROUGE score as the reward.
% {\bf DID NOT GET Also other evaluation metrics can be explored to determine which is best for this task or some metrics need to be defined like words before and after factoid answer position should be given higher weightage than the words copied directly from the question.} This new metric will lead to better evaluation of different methods. 
We plan on adding more variation to the data by annotating and correcting additional QA pairs both in factoid and existential questions.  

\bibliography{acl2020}
\bibliographystyle{acl_natbib}

\end{document}